\documentclass{article}

\usepackage{arxiv}

\usepackage[utf8]{inputenc} 
\usepackage[T1]{fontenc}    
\usepackage{hyperref}       
\usepackage{url}            
\usepackage{booktabs}       
\usepackage{amsfonts}       
\usepackage{nicefrac}       
\usepackage{microtype}      
\usepackage{lipsum}
\usepackage{graphicx}
\usepackage{algorithmic}
\usepackage{algorithm}
\usepackage{amsmath}

\title{Weakly Supervised Deep Learning Approach in Streaming Environments\thanks{This paper has been accepted for publication in The 2019 IEEE International Conference on Big Data (IEEE BigData 2019), Los Angeles, CA, USA.}}

\author{
  Mahardhika Pratama\thanks{Equal contribution.} \\
  School of Computer Science and Engineering\\
  Nanyang Technological University, Singapore \\
  \texttt{mpratama@ntu.edu.sg} \\
   \And
    Andri Ashfahani$^\dagger$ \\
  School of Computer Science and Engineering\\
  Nanyang Technological University, Singapore \\
  \texttt{andriash001@e.ntu.edu.sg} \\
  \And
 Mohamad Abdul Hady \\
  School of Computer Science and Engineering\\
  Nanyang Technological University, Singapore \\
  \texttt{hady.elits@gmail.com} \\
}

\begin{document}
\maketitle

\begin{abstract}
The feasibility of existing data stream algorithms is often hindered by the weakly supervised condition of data streams. A self-evolving deep neural network, namely Parsimonious Network (ParsNet), is proposed as a solution to various weakly-supervised data stream problems. A self-labelling strategy with hedge (SLASH) is proposed in which its auto-correction mechanism copes with \textit{the accumulation of mistakes} significantly affecting the model's generalization. ParsNet is developed from a closed-loop configuration of the self-evolving generative and discriminative training processes exploiting shared parameters in which its structure flexibly grows and shrinks to overcome the issue of concept drift with/without labels. The numerical evaluation has been performed under two challenging problems, namely sporadic access to ground truth and infinitely delayed access to the ground truth. Our numerical study shows the advantage of ParsNet with a substantial margin from its counterparts in the high-dimensional data streams and infinite delay simulation protocol. To support the reproducible research initiative, the source code of ParsNet along with supplementary materials are made available at \url{https://bit.ly/2qNW7p4}.
\end{abstract}


\section{Introduction}
Although incremental learning (IL) of data streams is well-established in the literature \cite{gamadatastream}, one of the major limitations in the existing approaches lies in their fully supervised nature in which the true class label has to become available immediately after receiving a data point. Some delay is expected in associating the target class to the incoming sample \cite{SCARGC}. In the realm of quality classification, the only way to understand the product quality of the manufacturing process is through visual inspection calling for the frequent stoppage of the machining process. The rate of delay is not constant, varying and often infinite. It is exemplified by the fact where the worn cases in the tool condition monitoring problem is often more difficult to be obtained than that of partially worn case. Damage of particular components of the aircraft engine often goes undetected because of the cost of fault analysis. These rationales outline urgent need of weakly-supervised data stream algorithms handling a variety of weakly-supervised learning situations regardless of the rate of delay and class distribution. The major challenge of weakly-supervised data streams is obvious in the fact that the concept drift might occur anytime with/without being accompanied by labels. 

One approach to reducing labelling cost of data streams makes use of the online active learning strategy which actively queries target labels of uncertain samples for model updates \cite{SAND,activelearning}. In \cite{SAND}, the confidence score is developed for sample selection. Although the active learning concept has been shown to reduce the labelling cost significantly \cite{activelearning}, these approaches work with the same assumption where the true class label can be immediately obtained regardless of the labelling cost. Another approach using the popular semi-supervised hashing algorithm is proposed and combined with the dynamic feature learning approach of auto-encoder \cite{ISLSD,IMM2012}. In \cite{DEVDAN}, the closed-loop configuration of the generative and discriminative processes are proposed to deal with partially labelled data streams. These algorithms are not compatible in the infinite delay scenario only having the ground truth access during the warm-up phase.

Another possible scenario for weakly-supervised data streams exists in the extreme latency problem. That is, labelled samples are provided only in the initial phase and true class labels never arrive afterward \cite{SCARGC}. This condition is more challenging to be tackled than the sporadic access to ground truth since it only involves a very small amount of labelled samples. COMPOSE is proposed as a solution to the infinite delay problem using the computational geometry approach called the $\alpha$ shape. As pointed out in \cite{MClassification}, COMPOSE suffers from high computational complexity. SCARGC is put forward in \cite{SCARGC} and exploits the pool-based approach. \cite{MClassification} adopts the self-evolving micro-clustering approach where unlabelled samples are associated with the most similar existing clusters. These approaches are derived from a non-deep learning strategy having the limited advantage in dealing with high input dimensions. 

A weakly-supervised deep learning approach, namely Parsimonious Network (ParsNet), is proposed and overcomes \textbf{both sporadic access and infinitely delayed access to ground truth}. The unique facet of ParsNet is seen in the combination of a self-evolving structure to handle the concept drift with/without label and \textbf{self-labelling with hedge} (SLASH) method addressing \textbf{the accumulation of mistakes} in generating pseudo labelled samples. The accumulation of mistakes is a major issue in the weakly-supervised data stream problem because noisy pseudo labels feed misleading information of the ideal decision boundary thus undermining the model's generalization. Note that ParsNet works in the one-scan fashion making the accumulation of mistakes difficult to handle. 

The SLASH functions as the weakly-supervised learning policy in which enrichment of class labels is performed with the auto-correction mechanism (hedge). It automatically associates unlabelled samples to the most confident class measured from the class posterior probability of ParsNet and Autonomous Gaussian Mixture Model (AGMM). Nonetheless, the self-labelling process risks on the accumulation of mistakes as a result of noisy pseudo labels. Our numerical study exhibits that a small quantity of noisy pseudo labels is sufficient to downgrade the model's predictive performance. The hedge implements the auto-correction mechanism automatically identifying noisy pseudo labels and preventing from forgetting important parameters. This strategy pushes the network parameters to its near optimal points. In a nutshell, the hedge protects ParsNet from performance deterioration due to wrong pseudo labels while accepting clean pseudo labels.  Augmentation of labelled samples is performed by injecting small noise to originally labelled samples thereby performing consistency regularization \cite{Mixmatch}. 

Adjustment of network width takes place in both generative and discriminative phases referring to reconstruction error and predictive error making possible for \textbf{identification of concept drifts with/without labels.} The hidden node evolution is governed by the advancement of network significance (NS) method \cite{ADL} relaxing the normal distribution assumption, adapting to the concept drifts and facilitating direct addition of $M$ hidden nodes to arrive at desirable network capacity quickly. This mechanism is underpinned by an autonomous Gaussian mixture model (AGMM) with a growing/pruning aptitude. That is, the approximation of probability density function $p(X)$ is obtained from dynamically evolved Gaussian components where a new Gaussian component is added if the concept drifts leading to $p(X)_t\neq p(X)_{t+1}$ ensues.

The major contribution of this paper is summed up in four facets: 1) online weakly-supervised deep neural network, ParsNet, is proposed to handle the problem of weakly-supervised data streams. ParsNet is applicable for both label's scarcity and extreme latency problems; 2) The SLASH is offered to deal with the scarcity of labelled samples via generation of pseudo label with protection against noisy pseudo labels; 3) Rapidly changing distributions of partially labelled data streams are dealt with a closed-loop configuration of flexible generative and discriminative phases having shared network parameters. Hidden units are automatically added/pruned in both phases thereby addressing the concept drift problem with/without target representation; 4) the advantage of ParsNet has been numerically validated in two challenging problems, \textbf{sporadic access to ground truth} and \textbf{infinitely delayed access to ground truth}. ParsNet demonstrates significant performance improvement over popular data stream algorithms in handling high-dimensional data streams with over 10\% performance gap while showing a similar pattern in the infinite delay case.

\section{Problem Formulation}
Weakly-supervised data stream problem is defined as a learning problem of sequentially arriving data batches $B_1,B_2,...,B_K$  where $K$ denotes the number of data batches that might not be bounded in practice. ParsNet adopts \textbf{a single-pass learning mode} without any epoch where data samples of $k$-th batch are directly discarded once learned. This scenario exemplifies ParsNet's feasibility in a strictly online learning procedure. Streaming data are captured without labels $B_k=X_k\in\Re^{N\times u}$ where $N,u$ denote the task size and the input dimension. The true class label $Y_k\in\Re^{N'}$ is solicited from particular labelling strategies. A weakly-supervised data stream learner is motivated by the prohibitive labelling cost or at least there exists some delay in obtaining the true class label thus leading to $N'<<N$. Note that the labelling delay is varying in nature and each target class incurs different labelling costs thereby causing sporadic access of true class labels. Another typical characteristic is observed in the presence of concept drifts $P(Y|X)_{t}\neq P(Y|X)_{t+1}$ \cite{gamadatastream} which might occur anytime with/without target labels.  

ParsNet is numerically validated under two simulation scenarios: \textbf{sporadic access to ground truth} and \textbf{infinitely delayed access to ground truth}. The first case outlines labelled samples $Y_k$ arrive sporadically without assuring the balanced class proportion. The target vector $Y_k$ only contains a subset of the original data batches having a much smaller size than the data batch $B_K$ where $N'<<N$. It hinders the application of active learning approach since the access of labelled samples other than those of $Y_k$ is not provided. This problem reflects different labelling costs of target classes and human factor such as boredom, fatigue, etc. The second case considers more complex case where a small amount of labelled samples $Y$ is only made available in the warm-up period while the remainder of data batches suffers from the absence of target classes. Moreover, the selection of initially labelled samples is taken without the assurance of the class distribution. That is, \textbf{the first batch of each problem} $B_1$ is chosen as the initially labelled samples without changing the original order. Initially labelled samples are made available once without being carried over to the next batches $B_2, B_3,...,B_K$. ParsNet only relies on augmented samples for model updates. The two procedures are simulated in \textbf{the prequential test-then-train protocol.}
\begin{figure}[!t]
\centerline{\includegraphics[scale=0.7]{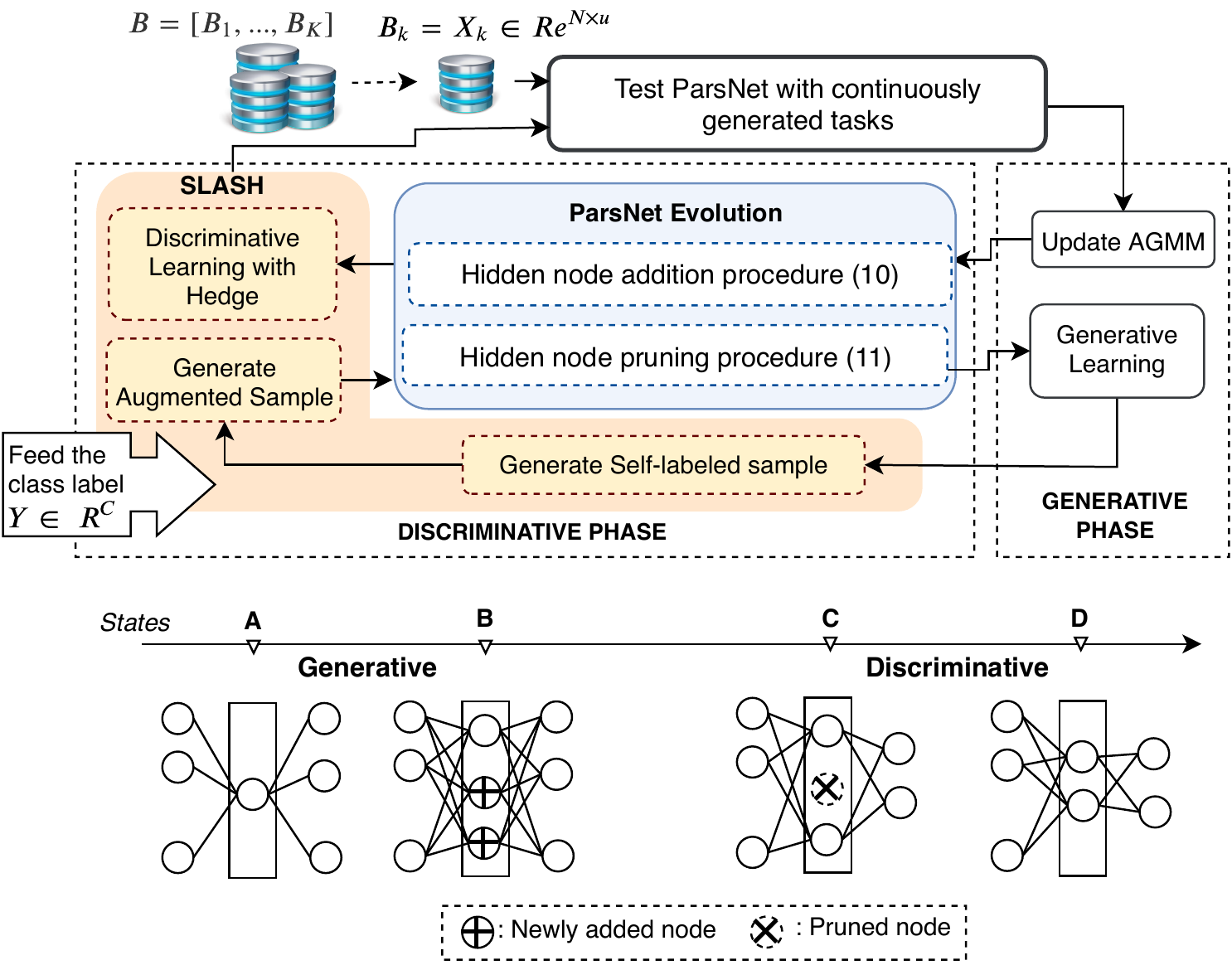}}
\caption{ParsNet learning policy and network evolution. \textbf{State A}: ParsNet starts its learning process with a hidden node in the generative phase. \textbf{State B}: Hidden nodes are added. \textbf{State C}: A hidden node is pruned on demand. \textbf{State D}: Final network structure.}
\label{ParsNetEvo}
\end{figure}

\section{ParsNet: Parsimonious Networks}
A complete picture of ParsNet learning policy is visualized in Fig. \ref{ParsNetEvo}. The learning process is initiated with the creation of AGMM which navigates the structural evolution: estimation of complex probability density function $p(X)$ and addition factor of hidden nodes $M$. AGMM features an open structure principle addressing the concept drift issue making the current estimation of probability density function to be obsolete $p(X)_{t}\neq p(X)_{t+1}$. Both ParsNet and AGMM start their learning process from scratch without a predefined structure. 

The generative phase takes place afterward and sets the initial condition of the discriminative phase. The generative phase also functions as an early alarm of changing learning conditions since adjustment of network width takes place in respect to the reconstruction error with the absence of ground truth. The discriminative step further evolves the generative network and runs the same structural learning phase with access to the ground truth. In other words, both phases utilize shared parameters. 

The weakly-supervised learning mechanism of ParsNet is governed by the idea of SLASH performing the self-labelling strategy with auto-correction. \textbf{The salient feature of the SLASH} lies in the solution of noisy pseudo labels uncharted in the existing literature. Note that the wrong pseudo label is a major problem due to what we call \textbf{the accumulation of mistakes}. Because of the absence of ground truth, mislabelled samples perturb the contour of decision boundary affecting the classification decision. This problem is complicated by the streaming environments because the access to old samples is prohibited. The auto-correction mechanism of the SLASH prevents ParsNet from forgetting its important parameters and automatically forces its parameters to its nearest optimal in the case of noisy pseudo labels. In addition, augmentation of originally labelled samples is performed to increase target representation.

The generative network is framed under the denoising auto-encoder (DAE) making use of partially corrupted input features via the masking noise. This step aims to extract robust features and to perform implicit regularization. As pointed out in \cite{learning_dynamic_AE}, the noise injecting mechanism in DAE supports faster convergence than the use of explicit regularization if a suitable noise level is selected. The generative network is expressed mathematically as follows:
\begin{equation}\label{generative}
    f_{enc}=s(\tilde{X}_n W_{enc}+b); \hat{X}_n=f_{dec}=s(f_{enc}W_{dec}+d)
\end{equation}
where $\tilde{X}_n\in\Re^{u}$ denotes partially destroyed input features $X_n$ by setting a subset of input features $u'$ to be blank. $W_{enc}\in\Re^{u\times R},b\in\Re^{R}$ are the connective weight and bias of encoder while $W_{dec}\in\Re^{R\times u},d\in\Re^{u}$ are the connective weight and bias of decoder. The tied-weight constraint is applied here and leads to $W_{dec}=W_{enc}'$. Here, $W_{enc},b$ are shared in the discriminative network as $W_{in},b$ and the softmax layer is inserted outputting the output posterior probability:
\begin{equation}\label{Discriminative}
    \hat{y}=softmax(s(X_n W_{in}+b_{in})W_{out}+c);%
\end{equation}
where $W_{out}\in\Re^{R\times C},c\in\Re^{C}$ are respectively the connective weight and bias of the soft-max layer. $C$ is the target classes.

\subsection{Network Evolution of ParsNet}
Growing and pruning processes occur in the generative and discriminative phases using an extension of the NS method.

\paragraph{Adjustment of network width.} The NS method \cite{ADL} is derived from the expectation of squared error $error^2$ under a normal distribution $p(X)$ which can be generalized to the popular bias-variance decomposition as per in (\ref{equation:ns}),  (\ref{equation:biasvar}).
\begin{eqnarray}
    NS=\int_{-\infty}^{\infty}(error^2)p(X)dX \label{equation:ns}\\
    NS=Var+(Bias)^2\label{equation:biasvar}
\end{eqnarray}
Note that the generative and discriminative learning phases are different only on the target variable inducing $error$. The reconstruction error, $(\hat{X}-X)$, is utilized in the generative phase, whereas the empirical error, $(\hat{y}-y)$, is utilized in the discriminative phase. (\ref{equation:biasvar}) can be expanded as follows:
\begin{eqnarray}
    NS_{d}=(E[\hat{y}^{2}]-E[\hat{y}]^{2})+(E[\hat{y}]-y)^{2}\label{nsdisc}\\
    NS_{g}=(E[\hat{X}^{2}]-E[\hat{X}]^{2})+(E[\hat{X}]-X)^{2}\label{nsgen}
\end{eqnarray}
where $NS_{d}$ and $NS_{g}$ denote the network significance of discriminative and generative phase, respectively. The NS method provides a statistical approximation of network's generalization power via the bias and variance decomposition to signify high bias and variance situations \cite{ADL}.

AGMM is proposed here to approximate the complex probability density function $p(X)$ and distinguishes itself from the original NS method \cite{ADL} derived from the unrealistic assumption of static Gaussian distribution. AGMM features a fully single-pass characteristic and an open structure with the growing and pruning processes of Gaussian components. A new component can be added if the concept drift is present. The use of AGMM leads to the estimation of the statistical  contribution of hidden nodes $E[s]$ as follows:
\begin{equation}
E[s]=\sum_{m=1}^{M}\int_{-\infty}^{\infty}s(W_{in} X+b_{in})N(X;\omega_m,c_m,\sigma_m)dX    
\end{equation}
where $N(X;\omega_m,c_m,\sigma_m)$ stands for the $m$-th GMM with the mixing coefficient $\omega_m$, center $c_m$, width $\sigma_m$. The integral can be solved independently for each GMM by following the finding of \cite{Murphy_Machine_Learning} where the sigmoid function can be approximated by the probit function $\Phi(\xi x)$ with $\xi=\frac{\pi}{8}$ and the integral of probit function is another probit function. The final expression of $E[s]$ is derived as follows:
\begin{equation}\label{expectationofh1}
    E[s]=\sum_{m=1}^{M}\omega_m s(\frac{c_m}{\sqrt{1+\pi\frac{\sigma_m^{2}}{8}}}W_{in}+b_{in})
\end{equation}
where $\sum_{m=1}^{M}\omega_m$ meets the partition of unity property. This approach not only summarizes distributional characteristic of already seen samples but also portrays network contribution navigated by the AGMM.

Consolidating the result from (\ref{expectationofh1}) enables us to calculate $Bias$ by substituting it into (\ref{nsdisc}) and (\ref{nsgen}). The high bias situation triggering the hidden node growing step is established:
\begin{equation}\label{HUgrow}
    \mu_{bias}^{n}+\sigma_{bias}^{n} \geq \mu_{bias}^{min}+\kappa\sigma_{bias}^{min}
\end{equation}
where $\mu_{bias}^{n}$, $\sigma_{bias}^{n}$ are respectively the empirical mean and standard deviation of $Bias$ while $\mu_{Bias}^{min},\sigma_{Bias}^{min}$ are the minimum bias up to the $n-th$ observation. Here, $\mu_{bias}^{min}$, $\sigma_{bias}^{min}$ are reset once (\ref{HUgrow}) is satisfied, whereas $\mu_{bias}^{n}$, $\sigma_{bias}^{n}$ are not reset because our numerical study exhibits performance degradation. That is, the estimation of the network's bias is reliable when considering all observations. (\ref{HUgrow}) is motivated by the sigma rule \cite{gamadatastream} where the confidence level depends on the choice of $\kappa$. To correct this shortcoming, $\kappa$ is set to be dynamic as $1.2exp(-Bias^{2})+0.8$. It enables to achieve $\kappa\in[1,2]$ leading to the confidence degree of $68.2\%$ to $95.2\%$. Note that (\ref{HUgrow}) has inherent drift detection capability because it is formulated from the statistical process control method \cite{gamadatastream}. 

If the condition (\ref{HUgrow}) is satisfied, $M$ number of hidden nodes are added simultaneously and are initialized using Xavier initialization. $M$ refers to the number of Gaussian components. This approach is one step ahead of existing approaches adopting one-by-one addition of hidden nodes considered too slow in adapting to concept drifts notably if a network structure has to be constructed from scratch \cite{ADL}. This strategy is supported by the nature of AGMM exploring the true complexity of data distributions in an online manner. 

ParsNet monitors the network's variance where complexity reduction procedure is carried out to cope with the high variance situation implying an over-complex network structure - prone to the over-fitting case. $Var$ can be obtained from (\ref{nsdisc}) and (\ref{nsgen}) by first calculating $\{E[\hat{X}^2],E[\hat{y}^2]\}$ and $\{E[\hat{X}],E[\hat{y}]\}$. Note that the first term is assumed to hold the i.i.d condition meaning that $E[s^2]=E[s]*E[s]$. The high variance condition is derived as follows:
\begin{equation} \label{HUprung}
    \mu_{var}^{n}+\sigma_{var}^{n}\geq \mu_{var}^{min}+2*\xi\sigma_{var}^{min}   
\end{equation}
where a reset scenario is applied to $\mu_{var}^{min}$, $\sigma_{var}^{min}$ if (\ref{HUprung}) is come across. It is observed that (\ref{HUprung}) is similar to (\ref{HUgrow}) except with the presence of $2\xi$ meant to avoid the direct-pruning-after-adding situation and the use of variance factor $\xi=1.2exp(-Var)+0.8$. Once the high variance condition is met, the statistical contribution of $r-th$ hidden node $E[s_r]$ is examined as per (\ref{expectationofh1}). Hidden nodes are deemed inactive and thus pruned without loss of generalization if its statistical contribution falls into the following condition: 
\begin{equation}
 E[s]_r\leq \mu_{E[s]}-\frac{1}{2}\sigma_{E[s]}, \quad \forall r = 1,\dots,R
\end{equation}
where $\mu_{E[s]}$, $\sigma_{E[s]}$ stand for the mean and standard deviation of hidden node statistical contributions. Note that this condition opens the likelihood for more than one hidden node to be pruned at once. This trait is important to induce sufficient reduction of network variance.

\paragraph{Autonomous Gaussian Mixture Model.} AGMM is utilized here to determine the probability density function $p(X)$ thereby relaxing from the normal distribution assumption. Unlike conventional GMM, AGMM features an open structure and operates in the one-pass mode. That is, AGMM parameters are not only tuned in the online fashion but also the growing and pruning scenarios are demonstrated to track any variation of data streams notably if $p(X)_t\neq p(X)_{t+1}$.

The growing process is governed by the Gaussian clustering method where a new component is incorporated to handle uncovered input space region as follows:
\begin{eqnarray}\label{GMMadd}
	\max_{m=1,...,M}\min_{j=1,...,u}\exp(-\frac{(X-c_{j,m})^{2}}{2\sigma_{j,m}^{2}})<\\ \nonumber \exp(-\frac{u \times \kappa}{4-2\exp(-u/20)})
\end{eqnarray}
This growing condition enumerates the spatial proximity of the most adjacent component to the incoming sample $x$ where the $min$ operator is utilized here to deal with unstructured problems: images, texts, etc. possessing a high input dimension. Furthermore, no gradient needs to be calculated making the use of $min$ operator feasible. The threshold on the right side is designed under the assumption that the majority of data points lies in $c_{j,m}+\kappa\sigma_{j,m}$ and beyond that it is considered abnormal samples. $u$ is inserted here to take into account the effect of input dimension. Moreover, $\kappa$ here enables the confidence level of sigma rule to be adjusted as the level of network bias, thereby relaxing too strict normal distribution case. 

The use of (\ref{GMMadd}) in the growing process often leads to an explosion of Gaussian components due to low variance direction of data distributions. A vigilance test is incorporated to control the size of AGMM components using the concept of available space. A new Gaussian component is added only if the most adjacent component does not have any available space to expand its size. That is, the tuning process of winning component incurs significant overlapping to other components. The concept of available space is realized by comparing the coverage span of winning component to all components. This strategy is also meant to prevent too general Gaussian components which lose its relevance to portray particular concepts. The vigilance test is formalized as follows:
\begin{align}
    \hat{V}_{win} &\geq \rho \label{vigilancetest} \sum_{m=1}^{M}\hat{V}, \quad m \neq win\\
    \rho &= \frac{\sum_{m=1}^M\sum_{j=1}^u score_{j,m}}{(M-1)\times u}, \quad m \neq win\label{vigilance}\\
    score_{j,m} &=\begin{cases}
0, & c_{j,win} - \sigma_{j,win} \leq c_{j,m} \leq c_{j,win} + \sigma_{j,win}\\
1, & otherwise
\end{cases} \nonumber
\end{align}
where $\rho$ denotes the vigilance test parameter. In this work, we adopt an adaptive vigilance test parameter $\rho$ adapting to the overlapping degree of winning Gaussian component to other components. It is designed as the portion of all components $c_m$ located outside $c_{win} \pm \sigma_{win}$ per input dimension as (\ref{vigilance}). $score_{j,m}=0$ supports the addition of new component whereas $score_{j,m}=1$ allows the winning Gaussian to expand its size. Finally, the growing mechanism in (\ref{GMMadd}) and (\ref{vigilancetest}) are combined with \textbf{AND} operator to determine the introduction of a new Gaussian component. The new component is crafted by simply setting the sample of interest $X$ as the centre of interest $c_{M+1}=X$ while the spread $\sigma_{M+1}$ is assigned with a small constant value $\alpha_4$.

The tuning scenario is carried out if the growing condition, (\ref{GMMadd}) \textbf{AND} (\ref{vigilancetest}), are violated. This step is meant to attain a fine-grained coverage of GMM in the input space where the winning GMM having the closest proximity to an incoming sample \cite{BARTFIS} is adjusted as follows:
\begin{eqnarray}
    c_{j,m}=c_{j,m}+\frac{X-c_{j,m}}{Sup_m+1},\nonumber\\
    \sigma_{j,m}^2=\sigma_{j,m}^2+\frac{(X-c_{j,m}^{2})-\sigma_{j,m}^{2}}{Sup_m+1},\nonumber\\ 
    Sup_m=Sup_m+1, \thinspace \thinspace Sup \in \Re^{M}
\end{eqnarray}
where $Sup_m$ denotes the support or population of $m-th$ GMM. An incoming sample is associated to the winning GMM while increasing the support of $m-th$ GMM. This step assures the convergence of the tuning scenario as the increase of GMM's population. The mixing coefficient, $\omega_m$, is obtained through the posterior probability of the GMM satisfying the partition of unity as follows \cite{BARTFIS}:
\begin{equation}
    \omega_m=P(N_m|X)=\frac{P(X|N_m)P(N_m)}{\sum_{m=1}^{M}P(X|N_m)P(N_m)}
\end{equation}
where $P(X|N_m)$ is the likelihood function arranged as $N(x|c_m,\sigma_m)$ while the prior probability, $P(N_m)$, is set as $\frac{Sup_m}{\sum_{m=1}^{M}Sup_m}$ or the relative support of $m-th$ component \cite{BARTFIS}.

The pruning procedure is implemented in the AGMM to get rid of inconsequential hidden units which play little during their lifespans or lose their relevance due to rapidly changing environments. Suppose that $\phi_m=\min_{j=1,..,u}\exp(-\frac{(X-c_{j,m})^{2}}{2\sigma_{j,m}^{2}})$ defines the activation degree of $m-th$ GMM, the relevance of $m-th$ GMM is introduced as the average of its activity, $Act$, over its lifespan \cite{BARTFIS,utility}:
\begin{equation}
    Act_m=\frac{\sum_{n=1}^{Lifespan}{\phi_m}}{Lifespan}
\end{equation}
where $Lifespan$ denotes the time period of a Gaussian component since it is inserted. The hidden unit pruning procedure is controlled by:
\begin{equation}\label{GMMprune}
    Act_m\leq |\mu_{Act}-0.5*\sigma_{Act}|
\end{equation}
 (\ref{GMMprune}) follows the half sigma rule. Furthermore, no component is pruned during a certain evaluation period making sure a new cluster is given the opportunity to develop its shape. The self-evolving property of AGMM enables the use of $M$ in the growing process of ParsNet since it explores the complexity of data distributions.
 
\subsection{SLASH}
The Self-Labelling Strategy with Hedge (SLASH) governs the parameter learning strategy formally expressed as a joint optimization problem as follows:
\begin{align}\label{loss}
    L &= L(\hat{X},X)+L(\hat{y},y)+L(\hat{y}_{aug},y_{aug})+L(\hat{y}_{ps},y_{ps}) + \frac{1}{2}\alpha_3\Gamma(\theta-\theta^*)^2
\end{align}
where $L(\hat{X},X)$ is the loss function of generative phase, $L(\hat{y},y)$ is the loss function of the true class label, $L(\hat{y}_{aug},y_{aug})$ is the loss function of augmented class labels, $L(\hat{y}_{ps},y_{ps})$ is the loss function of pseudo-class labels. The augmented label is a variation of originally labelled samples via the introduction of external perturbation while the pseudo-labelled data points are generated from the self-labelling process. Furthermore, the self-labelling process is carried out solely to unlabelled samples. 

The last term $\frac{1}{2} \alpha_3 \Gamma(\theta-\theta^*)^2$ functions as the hedge performing auto-correction against the noisy pseudo label and adopts the Synaptic Intelligence (SI) concept \cite{SI} originally proposed to overcome the catastrophic forgetting problem in the continual learning domain. We extend this concept as the hedge mechanism correcting noisy pseudo label. $\alpha_3$ is a regularization variable controlling how much information to be accepted from pseudo labels while $\Gamma$ stands for the parameter importance indicator memorizing influential network parameters before receiving the pseudo-label-induced model update. $\theta=[W_{in},b_{in},W_{out},c]$ consists of all network parameters meaning that the hedge affects all network parameters. $\theta^{*}$ is the optimal network parameters induced by the last true class label. $\Gamma$ is a regularization term reflecting the parameter's importance. Note that the originally labelled samples $(X,y)$, the augmented samples $(X_{aug},y_{aug})$ and the pseudo-labelled samples $(X_{ps},y_{ps})$ are mixed thereby making possible for the hedging strategy to be seamlessly executed. Note that structural evolution takes place only in the case of $L(\hat{y},y)$ and $L(\hat{X},X)$ because augmented samples are simply variations of labelled samples while pseudo-labelled samples risks on noisy information leading to incorrect estimation of network bias and variance. Since (\ref{loss}) is formulated as an unconstrained optimization problem, the SGD with \textbf{no epoch} is carried out alternately. This setting is meant to demonstrate the feasibility of ParsNet under the most challenging condition of data stream processing.

\paragraph{Self-Labelling Scenario.} The self-labelling process is carried out by examining the confidence degree of ParsNet and AGMM in which a label is propagated if both of them show confident and consistent prediction:
\begin{equation}
\label{confidencelabel}
    P(Y|X)_{AGMM}^{norm}\geq \alpha_1 \thinspace \textbf{AND} \thinspace P(Y|X)_{net}^{norm} \geq \alpha_2 
\end{equation}
where the output posterior probability of AGMM, $P(Y|X)_{AGMM}$, is determined as \cite{BayesianARTMAP}  $P(y_o|X)_{AGMM}=\frac{\sum_{m=1}^{M}P(y_o|Ny_m)P(Ny_m)P(X|Ny_m)}{\sum_{o=1}^{C}\sum_{m=1}^{M}P(y_o|Ny_m)P(X|Ny_m)P(Ny_m)}$. $P(y_o|Ny_m) =\frac{Supy_{o,m}}{\sum_{o=1}^{C}Supy_{o,m}}$ stands for the class posterior probability and $Supy \in \Re^{M\times C}$ is the frequency count \cite{BayesianARTMAP}. $P(Y|X)_{net}^{norm}$ denotes the normalized output posterior probability of ParsNet simply obtained from the two most dominant output classes \cite{LughoferActiveLearning} as $P(Y|X)_{net}^{norm}=\frac{y_1}{y_1+y_2}$. The normalization renders the class-invariant output of softmax layer behaving similarly to the binary classification case where uncertain prediction is indicated by the output posterior probability being close to 0.5 ($P(Y|X)_{AGMM}^{norm}\simeq0.5$, $P(Y|X)_{net}^{norm}\simeq0.5$). This case is often associated with data samples falling nearby the decision boundary. $\alpha_1,\alpha_2$ are two confidence thresholds. Note that a pseudo label is assigned only if both AGMM and ParsNet output the same class label.

\paragraph{Learning with Hedge.} The notion of a hedge, $\frac{1}{2} \alpha_3 \Gamma(\theta-\theta^*)^2$, is designed to prevent the performance drop due to the noisy pseudo label produced by the self-annotation process. It brings deviated parameters due to the wrong pseudo label back to their closest optimum value guided by the parameter importance indicator $\Gamma$. This mechanism is triggered before learning pseudo-labelled samples and controls the influence of pseudo-labelled samples to model updates. In other words, the regularization approach protects ParsNet against noisy pseudo labels by allowing $\theta$ to move from its optimal values $\theta^*$ only if a reliable pseudo label is fed by the self-labelling process. We utilize reconstruction error to determine the regularization factor $e_{recons}=L(\hat{X},X)=\frac{1}{2}\sum_u(X_u-\hat{X}_u)^2$ where the $Z$ score is used to scale the reconstruction error to the range of $[0,1]$ as per $\alpha_3=({e_{recons}-e_{recons}^{min}})/({e_{recons}^{max}-e_{recons}^{min}})$. This strategy implies the proportional reduction of the model update as the increase of reconstruction error. That is, noisy pseudo label distracts direction of gradients directly influencing the network's reconstruction power. 
As with the SI method \cite{SI}, $\Gamma$ plays an important role to steer ParsNet's parameters to its closest optimum parameters since it memorizes important network parameters. It is defined:
\begin{equation}\label{importance}
    \Gamma=\frac{\sum_{t=1}^{step}\Delta\theta\frac{\partial L}{\partial \theta}}{(\theta_T)^{2}+\epsilon}
\end{equation}
Since the hedge functions as the compensation of self-labelling mechanism, $\Gamma$ is updated only when observing the originally labelled and augmented samples to adjust the optimal parameters while being frozen when observing the pseudo label. Updating $\Gamma$ in the case of $L(\hat{y}_{aug},y_{aug})$ aims to further reinforce the memorization process since in this case, ParsNet has access to the ground truth. $\theta_T$ is the total parameter movement during the training process. $step$ denotes the total originally labelled and augmented samples received thus far while $\Delta \theta$ refers to the parameter's movement between two samples $\theta_t-\theta_{t-1}$ and $\frac{\partial L}{\partial \theta}$ is the gradient. $\epsilon$ is a small constant to avoid division with zero. The regularization via Hedge adjusts network parameters proportionally as the change of network parameters and the network gradient. In practice, the normalization is applied to $\Gamma$, $\Gamma_{new} = \frac{\Gamma_{old}}{||\Gamma_{old}||}$, as an effort to avoid exploding gradient. The underlying difference from the original version \cite{SI} lies in the use of regularization as a watchdog of the pseudo labelled samples where the transition between old and new parameters are unclear.

\paragraph{Generation of Augmented Samples.} Label enrichment scenario is also performed from augmentation of originally labelled samples or known as the consistency regularization approach. Augmented samples are formed as variations of labelled samples produced by the noise injecting mechanism where random Gaussian noise with zero mean is used to produce a corrupted version of labelled samples. $N(0,0.001)$ is specifically used for non-image problems while the image problem applies $N(0,33)$ being equivalent to $13\%$ corruption of an image. Since augmented points are sampled from originally labelled samples, it is not subject to the SLASH mechanism rather the SLASH parameters are adjusted here. This mechanism is crucial in the infinite delay case because labeled samples can be only accessed during the warm-up period. ParsNet only relies on augmented samples and pseudo-labelled samples for the infinite delay problem.

\section{Experiments}
In this section, the effectiveness of ParsNet is examined in two challenging cases: \textbf{sporadic access to ground truth and infinitely delayed access to ground truth}. An ablation study is conducted to analyze each learning component.

\subsection{Sporadic Access to Ground Truth}
\paragraph{Dataset.} Sporadic access to ground truth arranges a simulation environment where continuously arriving data batches comprises partially labelled samples. Our numerical study considers two proportions of labelled samples, 25\% and 50\%, in each data batch where the target classes are randomly distributed without any treatment of class distributions. Six prominent data stream problems are put forward: Rotated MNIST, Permuted MNIST \cite{permuttedMNIST}, weather \cite{Learn++NSE}, SEA \cite{SEA}, Hyperplane \cite{MOA} and RFID localization (Courtesy of Dr. Huang Sheng, Singapore). All of them except the RFID problem characterize \textbf{the non-stationary properties}. The RFID problem, however, presents the multi-class problem. The properties of the six problems are outlined in Table \ref{properties}. 
\begin{table}[!t]
\caption{Properties of the dataset.}
\begin{centering}
\label{properties} \scalebox{0.8}{ %
\begin{tabular}{lrrrrr}
\toprule 
Dataset  & IA  & C  & DP  & Tasks  & Class Proportion (\%)\tabularnewline
\midrule 
R. MNIST  & 784  & 10  & 62K  & 62  & $10.1;11.4;9.9;10.3;9.5;9;10;10.3;10;9.5$\tabularnewline
P. MNIST  & 784  & 10  & 70K  & 70  & $10;11.2;10.2;10.1;9.8;8.9;9.7;10.3;9.9;9.9$\tabularnewline
Weather  & 8  & 2  & 18K  & 18  & $31.7;68.3$\tabularnewline
Sea  & 3  & 2  & 120K  & 120  & $63.15;36.85$\tabularnewline
Hyperplane  & 4  & 2  & 25K  & 25  & $49.83;50.17$\tabularnewline
RFID & 3 & 4 & 280K & 280 & $24.96;24.88;25.01;25.15$\tabularnewline
\bottomrule
\end{tabular}} 
\par\end{centering}
\vspace{5pt}
{IA: input attributes, C: classes, DP: data points.} 
\end{table}

\paragraph{Baseline and parameter Setting.} Five recently published algorithms, namely ADL \cite{ADL}, DEVDAN \cite{DEVDAN}, OMB \cite{OMB}, Learn++ (L++) \cite{polikar2001learn++} and Learn++NSE (LNSE) \cite{Learn++NSE} are considered as baselines. We execute each algorithm in the same simulation protocol and computational resources using \textbf{their published codes} to ensure a fair comparison. Their original hyper-parameter setting is retained. If their performance is surprisingly poor, they are tuned and their best results are reported here. On the other hand, the predefined parameters of ParsNet are selected as $\alpha_1=0.55$, $\alpha_2=0.6$ and $\alpha_4=0.1$ while learning rates of generative and discriminative phases are chosen respectively as $0.01$ and $0.001$. This setting is fixed for all numerical study in this paper and is obtained simply from minor hand-tuning mechanism to show the non-ad-hoc nature of ParsNet. \textbf{The source code of ParsNet along with all datasets and supplemental document is offered to support the reproducible research initiative}. The link is given in the abstract of this paper. Numerical results of other algorithms are produced using \textbf{50\% labelled samples}.

\paragraph{Simulation Protocol.} The prequential test-then-train protocol is followed here as per the guideline of \cite{datastreamevaluation}. A model is forced to predict the entire samples of each incoming task before utilizing them for model updates. Numerical results are obtained from the average of numerical results per data batch. The numerical results of ParsNet are taken from the average of five independent runs to arrive at conclusive learning performance. Numerical results are statistically validated using the Wilcoxon signed-rank test where ($^*$) indicates statistically significant differences. 

\paragraph{Numerical Results.} The advantage of ParsNet is shown in Table \ref{result-1-2-1-2} where ParsNet outperforms other algorithms in four of six datasets. Substantial performance improvements are attained by ParsNet in the rotated MNIST and permuted MNIST problems with over 10\% margin. Note that the numerical results of Lean++ and Learn++NSE  cannot be produced in the two problems because they do not scale well with high-dimensional data streams. This result signifies the efficacy of the SLASH helping to achieve major performance improvement in the complex problem having a high input dimension. The self-labelling mechanism of SLASH enriches labelled samples while the hedging mechanism prevents the accumulation of mistakes. The hedge rejects noisy pseudo labels by preventing the network parameters to be distracted too far from their previous optimal values.
\begin{table}[!t]
\caption{The numerical results of sporadic access to ground truth scenario.}
\begin{centering}
\label{result-1-2-1-2} \scalebox{0.8}{ %
\begin{tabular}{lrrrrrrrr}
\toprule 
 &  & {ADL} & {OMB} & {L++} & {L++NSE} & {DEVDAN} & \multicolumn{2}{c}{ParsNet}\tabularnewline
 &  &  &  &  &  &  & $50\%$  & $25\%$ \tabularnewline
\midrule 
Rotated MNIST & CR  & $42.86$  & $26$  & N/A  & N/A  & $42.33$  & $\textbf{64.32}\pm\textbf{0.87}^{*}$  & $58\pm0.91$\tabularnewline
 & TrT  & $0.26$  & N/A  & N/A  & N/A  & $2.34$  & $23.59$  & $12.11$\tabularnewline
 & HN  & $10$  & N/A  & N/A  & N/A  & $27.1$  & $291.7$  & $192.3$\tabularnewline
 & HL  & $1$  & $3$  & N/A  & N/A  & $1$  & $1$  & $1$\tabularnewline
 & PS  & N/A  & N/A  & N/A  & N/A  & N/A  & $23.6$  & $35.4$\tabularnewline
\midrule 
Permuted MNIST & CR  & $76.86$  & $11.22$  & N/A  & N/A  & $73.07$  & $\textbf{83.91}\pm\textbf{0.54}^{*}$  & $80.63\pm0.64$ \tabularnewline
 & TrT  & $0.29$  & N/A  & N/A  & N/A  & $2.4$  & $26.45$  & $14.78$\tabularnewline
 & HN  & $27.3$  & N/A  & N/A  & N/A  & $54.5$  & $363.1$  & $216.8$\tabularnewline
 & HL  & $1.01$  & $3$  & N/A  & N/A  & $1$  & $1$  & $1$\tabularnewline
 & PS  & N/A  & N/A  & N/A  & N/A  & N/A  & $77.63$  & $99.9$\tabularnewline
\midrule 
Weather & CR  & $71.63$  & $65.32$  & $\textbf{75.06}$  & $74.01$  & $70.75$  & $72.58\pm1.4{}^{*}$  & $70.24\pm1.19$\tabularnewline
 & TrT  & $0.05$  & N/A  & $1.66$  & $0.47$  & $0.6$  & $0.66$  & $0.55$\tabularnewline
 & HN  & $8.1$  & N/A  & N/A  & N/A  & $10.3$  & $15.1$  & $13.5$\tabularnewline
 & HL  & $1$  & $2$  & $2$  & $19$  & $1$  & $1$  & $1$\tabularnewline
 & PS  & N/A  & N/A  & N/A  & N/A  & N/A  & $147.6$  & $271.4$\tabularnewline
\midrule 
SEA & CR  & $91.22$  & $87.86$  & $87.62$  & $90.01$  & $91.31$  & $\textbf{91.41}\pm\textbf{0.18}^{*}$  & $89.47\pm0.47$\tabularnewline
 & TrT  & $0.07$  & N/A  & $9.88$  & $2.98$  & $0.6$  & $0.78$  & $0.72$\tabularnewline
 & HN  & $19.8$  & N/A  & N/A  & N/A  & $21.5$  & $57.4$  & $68.3$\tabularnewline
 & HL  & $1$  & $2$  & $2$  & $100$  & $1$  & $1$  & $1$\tabularnewline
 & PS  & N/A  & N/A  & N/A  & N/A  & N/A  & $292.5$  & $453.6$\tabularnewline
\midrule 
Hyperplane & CR  & $91.77$  & $86.78$  & $88.61$  & $90.06$  & $91.47$  & $\textbf{92.28}\pm\textbf{0.24}$  & $91.67\pm0.06$\tabularnewline
 & TrT  & $0.05$  & N/A  & $10.98$  & $4.31$  & $0.58$  & $0.49$  & $0.43$\tabularnewline
 & HN  & $3.2$  & N/A  & N/A  & N/A  & $23.5$  & $8.3$  & $8.6$\tabularnewline
 & HL  & $1$  & $2$  & $2$  & $120$  & $1$  & $1$  & $1$\tabularnewline
 & PS  & N/A  & N/A  & N/A  & N/A  & N/A  & $0$  & $0$\tabularnewline
\midrule 
RFID & CR  & $98.66$  & $99$  & $91.7$  & $\textbf{99.53}$  & $98.56$  & $98.89\pm0.1^{*}$  & $98.46\pm0.1$\tabularnewline
 & TrT  & $0.05$  & N/A  & $22.41$  & $8.73$  & $0.64$  & $1.49$  & $1.03$\tabularnewline
 & HN  & $19.9$  & N/A  & N/A  & N/A  & $48.2$  & $191.4$  & $166.7$\tabularnewline
 & HL  & $1$  & $2$  & $4$  & $280$  & $1$  & $1$  & $1$\tabularnewline
 & PS  & N/A  & N/A  & N/A  & N/A  & N/A  & $0$  & $0$\tabularnewline
\bottomrule
\end{tabular}} 
\par\end{centering}
\vspace{5pt}
 {CR: classification rate, TrT: training time, HN: hidden nodes, HL: hidden layers, PS: Number of pseudo labels. $^*$: indicates that the numerical results of ParsNet and other methods are significantly different.} 
\end{table}

This observation also has a strong correlation to the nature of the image classification problem where the noise injecting mechanism in the data augmentation process is capable of increasing the number of labelled samples without suffering from the consistency issue. In short, the noise injecting mechanism does not affect the class densities. ParsNet also produces better accuracy in the hyperplane and SEA problems with statistically significant differences. Although ParsNet is behind Learn++ in the weather problem, it is mainly attributed to the NN's characteristic which does not cope well with the high uncertainties. A similar situation is also observed in both DEVDAN and ADL. ParsNet is also less accurate than Learn++NSE in the RFID problems but still better than other algorithms. In the realm of execution time, ParsNet incurs a moderate increase of computational overhead from DEVDAN and ADL but remains much faster than Learn++NSE and Learn++. This observation is not surprising due to the additional training steps of ParsNet compared to ADL and DEVDAN. Another important finding is shown in Table \ref{result-1-2-1-1-2} reporting the precision and recall of ParsNet produced using 50\% lebelled samples. It is depicted that precision and recall exhibit a relatively small gap meaning that ParsNet's classification decision is not biased to one of the classes.
\begin{table}[!t]
\begin{centering}
\label{precisionrecall} \caption{Precision and recall of sporadic access to ground truth scenario.}
\centering{}\label{result-1-2-1-1-2} \scalebox{0.8}{ %
\begin{tabular}{llr}
\toprule 
{Rotated MNIST } & P & $[0.78;0.9;0.61;0.64;0.51;0.62;0.58;0.68;0.59;0.52]$\tabularnewline
 & R & $[0.82;0.94;0.57;0.62;0.54;0.56;0.57;0.68;0.62;0.53]$\tabularnewline
\midrule 
{Permuted MNIST } & P & $[0.92;0.88;0.86;0.83;0.85;0.72;0.90;0.84;0.82;0.80]$\tabularnewline
 & R & $[0.89;0.93;0.82;0.81;0.83;0.80;0.89;0.86;0.79;0.79]$\tabularnewline
\midrule 
{Weather} & P & $0.58$\tabularnewline 
 & R & $0.49$\tabularnewline 
\midrule 
{SEA} & P & $0.92$\tabularnewline
 & R & $0.85$\tabularnewline 
\midrule 
{Hyperplane } & P & $0.92$\tabularnewline 
 & R & $0.92$\tabularnewline 
\midrule 
{RFID} & P & $[0.9998;0.9884;0.98;0.99]$\tabularnewline
 & R & $[1;0.99;0.98;0.99]$\tabularnewline
\bottomrule
\end{tabular}} 
\par\end{centering}
\vspace{5pt}
 {P: Precision. R: Recall.} 
\end{table}

\subsection{Infinitely Delayed Access to Ground Truth}
\paragraph{Simulation Protocol.} The second numerical study is carried out by following the infinite delay protocol where the true class label is only provided during the warm-up phase while the remainder of entire streams arrive without labels. Our numerical study makes use of the same datasets and procedure as the first numerical study except only for that \textbf{the first data batch} is used in the warm-up phase without changing the original data order thus leading to the very small amount of labeled samples. The ratio of labelled samples and all samples, as well as the class proportion, are provided in Table \ref{result-1-2-1-1-1}.
\begin{table}[!t]
\begin{centering}
\label{infinitedelay-1} \caption{The numerical results of infinitely delayed access to ground truth scenario.}
\centering{}\label{result-1-2-1-1-1} \scalebox{0.8}{ %
\begin{tabular}{lrrrrrr}
\toprule 
 & R. MNIST  & P. MNIST  & Weather & SEA & Hyperplane & RFID\tabularnewline
\midrule 
ParsNet  & $\textbf{48.46}\pm\textbf{5.96}^{*}$  & $\textbf{46.25}\pm\textbf{1.11}^{*}$  & $\textbf{72.33}\pm\textbf{0.57}^{*}$  & $\textbf{88.01}\pm\textbf{0.34}^{*}$  & $\textbf{86.82}\pm\textbf{0.24}^{*}$  & $\textbf{94.17}\pm\textbf{1.09}^{*}$\tabularnewline
S-1NN & $23.81$ & $33.31$ & $69.89$ & $78.12$ & $78.20$ & $59.13$\tabularnewline
S-SVM & $20.92$ & $33.02$ & $71.62$ & $82.59$ & $81.89$ & $52.11$\tabularnewline
DEVDAN & $28.98$ & $36.65$ & $70.79$ & $78.49$ & $62.56$ & $58.2$\tabularnewline
\midrule
NLS/NS & $0.02$ & $0.014$ & $0.06$ & $0.01$ & $0.008$ & $0.004$\tabularnewline
Class Proportion (\%) & $A$ & $B$ & $32.7;67.3$ & $63.2;36.8$ & $47.8;52.2$ & $C$\tabularnewline
\bottomrule
\end{tabular}}
\par\end{centering}
\vspace{5pt}
 {${\scriptstyle A=9;10.7;10.3;9.8;11.4;9;9;11.4;8.8;10.6}$.\\ ${\scriptstyle B=7.4;12.1;10.5;10;9.4;9.8;10.3;11;9.7;9.8}$. ${\scriptstyle C=26.4;26.3;21;26.4}$.\\ NS: Number of all samples. NLS: Number of labelled samples.} 
\end{table}

\paragraph{Baseline.} SCARGC \cite{SCARGC} is put forward as the baseline because it is specifically designed for the infinite delay problem and considered as a state-of-the-art algorithm. It has been reported to be better than other algorithms such as COMPOSE \cite{COMPOSE}. Two versions of SCARGC built upon SVM (S-SVM) and 1NN (S-1NN) are used here where their results are obtained using \textbf{their published codes} under the same computational environment as ParsNet. Besides, DEVDAN is added as our baseline because it has the generative phase to handle the infinite delay situation. Table \ref{result-1-2-1-1-1} reports numerical results.

\paragraph{Numerical Results.} Our numerical results in Table \ref{result-1-2-1-1-1} confirm the advantage of ParsNet in the infinite delay scenario. ParsNet beats the three baselines with conclusive performance differences - statistically significant in all cases. It is worth mentioning that the infinite delay problem is more challenging than the sporadic access to ground truth due to a very low proportion of labelled samples. Both ParsNet and DEVDAN suffers from performance drop compared to the first case.

\subsection{Ablation Study}
The ablation study covers three configurations: (A) without AGMM, (B) without network evolution, (C) without SLASH. Our ablation study is carried out under the sporadic access to ground truth. Furthermore, the learning performance is studied under $25\%$ labelled samples using the rotated MNIST problem. Numerical results are reported in Table \ref{ablationstudy}. It is observed from configuration (A) that the deactivation of AGMM significantly compromises the training performance because it becomes too slow to achieve desirable network capacity and to adapt to the concept drift. Poor performance is perceived in the configuration (B) if the structural learning mechanism is switched off. The absence of SLASH in configuration C contributes to around 3\% degradation on accuracy. The concept of self-evolving structure has been studied intensively in the context of fuzzy and RBF networks \cite{EFSsurvey}.
\begin{table}[!t]
\begin{centering}
\caption{The numerical result of ablation study.}
\centering{}\label{result-1-1} \scalebox{0.8}{ %
\begin{tabular}{lrrrrr}
\toprule 
 &  & ParsNet  & A  & B  & C \tabularnewline
\midrule 
Rotated MNIST & CR  & $\textbf{58}\pm\textbf{0.91}$  & $41.78\pm1.5$  & $41.32\pm0.78$  & $55.71\pm0.4$\tabularnewline
 & TrT  & $12.11$  & $1.66$  & $3.3$  & $10.29$\tabularnewline
 & HN  & $192.3$  & $13.19$  & $10$  & $162.6$\tabularnewline
 & PS  & $35.4$  & $0$  & $29.5$  & $0$\tabularnewline
\bottomrule
\end{tabular}} \label{ablationstudy}
\par\end{centering}
\end{table}

\section{Conclusions}
A weakly-supervised deep learning algorithm, ParsNet, is proposed for handling various weakly supervised data stream problems. Our numerical study under a scarcity of labelled samples as well as extreme label latency demonstrates the efficacy of ParsNet. The major performance improvement by over 10\% is attained in handling high-dimensional data streams under the minor supervisory mechanisms and in dealing with the infinite delay scenario. 

\section*{Acknowledgement}
This paper is financially supported by National Research Foundation with its Industry Alignment Fund Pre-Positioning Programme in AME domain [AwardNo.: A19C1a0018].

\bibliographystyle{unsrt}  
 \bibliography{references}
\end{document}